\documentclass[preprint, 12pt]{elsarticle}

\makeatletter
\def\ps@pprintTitle{%
  \let\@oddhead\@empty
  \let\@evenhead\@empty
  \let\@oddfoot\@empty
  \let\@evenfoot\@oddfoot
}
\makeatother

\usepackage{amssymb}
\usepackage{times}
\usepackage{soul}
\usepackage{url}
\usepackage{graphicx}
\usepackage{amsmath}
\usepackage{amsthm}
\usepackage{booktabs}
\usepackage{algorithm}
\usepackage{algorithmic}
\usepackage{comment}
\usepackage{subfigure}
\usepackage{multirow}
\usepackage{cite}
\usepackage{lettrine}
\usepackage{color}
\usepackage[hidelinks]{hyperref}
\usepackage{xcolor}

\begin{document}
\sloppy

\begin{frontmatter}



\title{Focalized Contrastive View-invariant Learning for Self-supervised Skeleton-based Action Recognition}


\author[a,b]{Qianhui Men\corref{*}\fnref{1}}
\emailauthor{qianhui.men@eng.ox.ac.uk}{Q. Men}
\author[c]{Edmond S. L. Ho}
\author[d]{Hubert P. H. Shum}
\author[a]{Howard Leung}

\cortext[*]{Corresponding author}
\fntext[1]{Present address: Department of Engineering Science, University of Oxford, OX1 3PJ, UK. Work done while Qianhui Men was at City University of Hong Kong.}

\address[a]{Department of Computer Science, City University of Hong Kong, Hong Kong SAR, China}
\address[b]{Department of Engineering Science, University of Oxford, Oxford, OX1 3PJ, United Kingdom}
\address[c]{School of Computing Science, University of Glasgow, Glasgow, G12 8RZ, United Kingdom}
\address[d]{Department of Computer Science, Durham University, Durham, DH1 3LE, United Kingdom}

\begin{abstract}

Learning view-invariant representation is a key to improving feature discrimination power for skeleton-based action recognition. Existing approaches cannot effectively remove the impact of viewpoint due to the implicit view-dependent representations. In this work, we propose a self-supervised framework called Focalized Contrastive View-invariant Learning (FoCoViL), which significantly suppresses the view-specific information on the representation space where the viewpoints are coarsely aligned. By maximizing mutual information with an effective contrastive loss between multi-view sample pairs, FoCoViL associates actions with common view-invariant properties and simultaneously separates the dissimilar ones. We further propose an adaptive focalization method based on pairwise similarity to enhance contrastive learning for a clearer cluster boundary in the learned space. Different from many existing self-supervised representation learning work that rely heavily on supervised classifiers, FoCoViL performs well on both unsupervised and supervised classifiers with superior recognition performance. Extensive experiments also show that the proposed contrastive-based focalization generates a more discriminative latent representation. 

\end{abstract}

\begin{keyword}


self-supervised learning \sep skeleton-based action recognition \sep contrastive learning
\end{keyword}

\end{frontmatter}


\section{Introduction}
Self-supervised skeletal human action recognition (HAR) aims at automatically detecting a robust representation to cluster and identify actions from class-agnostic skeletal data. Compared to supervised models that heavily rely on action labels \citep{yan2018spatial, zhang2019view, zhang2020semantics}, recognition without manual labeling is considered more efficient and more comprehensive to learn representative features with large-scale data. A few unsupervised attempts~\citep{zheng2018unsupervised,su2020predict,nie2020unsupervised,yang2021skeleton,guo2022contrastive} have recently achieved classification results comparable to supervised models, which indicates that label information may not be necessary for extracting useful representations for discriminating action dynamics. In this work, we consider the challenging domain of self-supervised action recognition from multi-view features, where the action sequences are captured under different viewpoints. The diverse view appearance introduces large intra-class variations in the feature representation that substantially impacts the clustering performance. \textcolor{black}{Unlike supervised view-invariant learning that benefits from action labels, learning without label guidance is more challenging that usually requires detecting implicit consistency between viewpoints.}

\textcolor{black}{Existing skeleton-based HAR works learn view-invariant features from the skeleton descriptions enriched by multi-view observations~\citep{zhang2019view,nie2020unsupervised} or unseen viewpoints~\citep{rahmani2017learning}.} A simple yet effective pre-processing scheme is to align the body key joints with a local coordinate system~\citep{liu2017enhanced,lee2017ensemble}. However, since this view-invariant transformation is sensitive to the quality of the posture captured from different viewpoints, such as different levels of self-occlusions, the transformed multi-view actions are still mismatched with many inherent view-specific representations~\citep{nie2019view}. Later on, deep neural networks are utilized to automatically search for the optimal viewpoints for every skeleton sequence~\citep{zhang2019view, liu2021adaptive}, which requires strong supervision to guide this additional training. In unsupervised learning, an adversary view-aware classifier is introduced in~\citep{li2018unsupervised} to discard view information from RGB and depth data. Another attempt~\citep{nie2020unsupervised} learns the view-variant and view-invariant features from spatial and temporal skeletal representations respectively, while the recognition performance is less satisfactory on the multi-view actions. So far, removing the viewpoint impact in the self-supervised skeleton recognition is still an open problem.

\begin{figure}
    \centering
    \includegraphics[width=1.0\columnwidth]{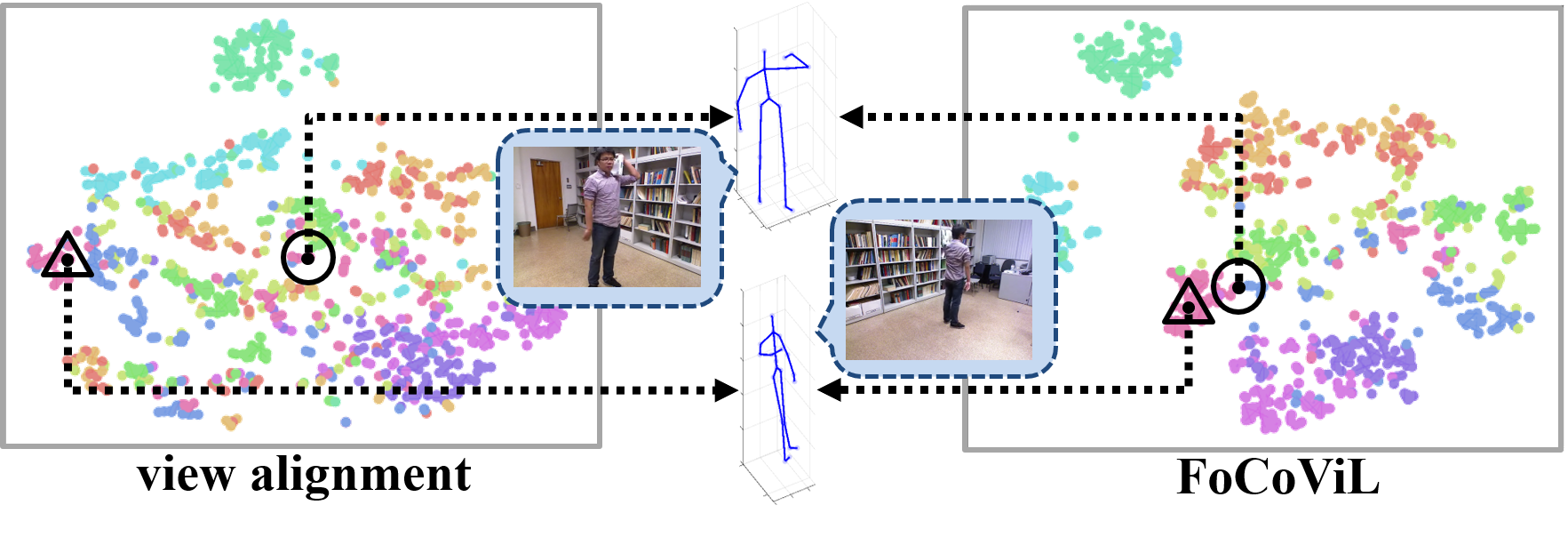}
    \caption{Visualizing the latent space when comparing an action pair of \textit{carrying} from the same scene but different viewpoints. Different colors in the t-SNE visualization refer to the attributes of real classes. The view alignment cannot handle the inherent heterogeneity between viewpoints, such as different levels of self-occlusions shown in the middle skeletons, yielding a less satisfactory space distribution. With FoCoViL, the actions under the same scene but different viewpoints (highlighted in circles and triangles) are geometrically closer in the latent space, which achieves better view invariance.}
    \label{fig:view_invariant}
\end{figure}

In this paper, we propose FoCoViL, the focalized contrastive view-invariant learning framework, for view-independent and discriminative self-supervised action recognition. FoCoViL consists of two complementary components, namely contrastive view-invariant learning (CoViL) and focalization. Figure~\ref{fig:view_invariant} shows the effect of FoCoViL on learning a view-invariant latent space, where the action representations of the same scene are more closely distributed compared to CoViL.

First, CoViL explores the implicit relationship between viewpoints. With the facing directions aligned~\citep{lee2017ensemble}, CoViL works as a refinement scheme to group the multi-view actions by maximizing their mutual information under different viewpoints, such that the learned representation is robust to view changes. Specifically, under a self-supervised auto-encoder backbone, we propose to maximize the agreement of the actions under the same scene but different viewpoints (\textit{i.e.} positive pairs ``$+$”), which helps extract the common features among them that are view-invariant. Meanwhile, we propose to enlarge the disagreement of actions under different scenes (\textit{i.e.} negative pairs ``$-$"), which benefits the clustering with a sparse latent space. The two goals are jointly achieved by a close form of contrastive loss~\citep{chen2020simple} for its superior ability to find and compare the similarity and dissimilarity in the self-supervised representations. With CoViL, we construct a latent space that is more robust to view dynamics compared to the one generated by the low-level viewpoint alignment~\citep{su2020predict}.

Second, we propose to enhance the latent space by wrapping a novel focalization method around contrastive learning in CoViL. This is to solve the imbalanced training data issue inherent in many existing self-supervised systems - the hard samples that dominate the misclassification are not fully investigated, leading to an ambiguous sample distribution in the latent space.
To mitigate imbalance in contrastive learning, several works that are highly related to ours mainly focus on mining hard negatives, such as synthesizing new samples~\citep{dai2019generative,NEURIPS2020_f7cade80} or using class labels as priors~\citep{kang2021exploring}. In contrast, our method considers adaptively ``focalizing" both the hard positives and negatives under the learned representative similarity. We take advantage of the effective pairwise similarity estimation in CoViL to dynamically identify and re-balance the easy and hard multi-view action pairs. This is done by defining the hard pairs as either sparse positive pairs (same scenes that are far away) or dense negative pairs (different scenes that are close) in the projected latent space, and the easy ones the other way round. The proposed focalization reduces the weightings of easy pairs that provided limited information while focusing on pushing hard negative pairs away and pulling hard positives closer, thereby enforcing a more distinct decision boundary in the latent space. 

Experimental results show that FoCoViL outperforms state-of-the-art self-supervised models on five benchmark 3D action datasets including Northwestern-UCLA (N-UCLA)~\citep{wang2014cross}, NTU RGB+D 60~\citep{shahroudy2016ntu}, NTU RGB+D 120~\citep{liu2019ntu}, UWA 3D Multiview Activity II (UWA3D)~\citep{rahmani2014hopc}, and PKU-MMD~\citep{liu2017pku}. \textcolor{black}{Unlike some self-supervised representation learning approaches~\citep{li20213d,guo2022contrastive} rely heavily on supervised classifiers, FoCoViL performs well with both supervised and unsupervised classifiers.} The extensive experiments on representation space evaluation also indicate that the proposed FoCoViL produces a more robust latent space.

The main contributions are summarized in three folds: 
\begin{itemize}
    \item We propose a self-supervised framework to progressively learn a discriminative skeleton-based action representation that is robust for both supervised and unsupervised evaluation protocols\footnote{The source code is publicly available at:     
    {\color{blue}\url{https://drive.google.com/file/d/1VKRF2S3-LrOiXV4BLSjxIlCUewMnxH_W}}.}.
    \item We propose contrastive view-invariant learning, which maximizes the mutual information between multi-view action pairs by adapting contrastive learning, aiming to refine the latent representations with high-level view-invariant features.
    \item \textcolor{black}{As a novel attempt of applying focalization to contrastive learning, we have demonstrated its feasibility of learning a more robust and unbiased representation with the action recognition task. }

\end{itemize}    

The rest of this paper is organized as follows. Section~\ref{sec:related_work} reviews the related background research. Section~\ref{sec:focovil} presents the proposed FoCoViL framework for self-supervised action recognition. The experiments and analysis are conducted in Section~\ref{sec:experiments} with quantitative recognition and reconstruction results, and qualitative latent space evaluations. Finally, we conclude this paper in Section~\ref{sec:con}.

\section{Related Work}
\label{sec:related_work}
\subsection{Self-supervised Action Representation Learning}
Learning of action representation has been proposed for many years in computer vision applications. \textcolor{black}{The learned latent representation usually includes action semantics which is feasible for multiple downstream tasks, such as action classification~\citep{ke2017new,tran2018closer,piergiovanni2020evolving} and motion generation~\citep{butepage2017deep,wang2019spatio,men2020quadruple}.} Among many self-supervised feature extraction baselines, the auto-encoder is usually adopted to learn the representation space with its superior ability to denoise the action information. Holden et al.~\citep{holden2015learning} first proposed a convolutional auto-encoder to construct the latent space from the encoder output. By operating the high-level features, the model is functional in many areas, such as action interpolation and comparison. Later on, with a hierarchical RNN auto-encoder in both spatial and temporal domains, Wang et al.~\citep{wang2019spatio} developed a high-quality representation space that is motivated for precise action modeling.

\textcolor{black}{As a vision-based learning task, the effectiveness of self-supervised representation is frequently explored in RGB-based action understanding. To learn the contextual coherence in action representation, Lai and Xie~\citep{lai2019self} matched pixel-wise correspondence from the spatial-temporal color information. Han et al.~\citep{han2020self} exploited action representations from multiple modalities of RGB streams and optical flow. However, these RGB-based action recognition models usually learn contrastive representation from background or visual consistency~\citep{wang2021removing}. With the visual information unavailable, the challenge of learning self-supervised skeleton-based action representation mainly comes from the diverse pose information under different view observations~\citep{nie2020unsupervised,li20213d}.} 

In self-supervised skeleton-based action representation learning, existing works such as~\citep{zheng2018unsupervised,demisse2018pose,kundu2019unsupervised,su2020predict} mainly focus on preserving the action-dependent features as much as possible to identify samples~\citep{yue2022action}. \textcolor{black}{For example, an adversarial discriminator is used to assist the auto-encoder to rectify the reconstructed action for a more distinctive representation~\citep{kundu2019unsupervised,zheng2018unsupervised}.} 
Since the encoder is dominant in disclosing the action features, Su et al.~\citep{su2020predict} proposed to strengthen the encoder by exploiting different auto-encoder structures (P\&C), such as fixing the encoded state or the decoder weights. Apart from an action-level auto-encoder, they also designed an additional feature-level auto-encoder to reduce the dimensionality of the learned representations, which results in a two-round training process. Because of the lack of communication within latent space, the derived action representation of these works is not robust to large intra-class variations. 
Other prior works considered learning feature representations by modeling actions with denoised poses (Denoised-LSTM~\citep{demisse2018pose}), different temporal patterns (MS$^2$L~\citep{lin2020ms2l}, MCAE-MP~\citep{xu2021unsupervised}), different spatial-temporal augmentations (AS-CAL~\citep{rao2021augmented}, ST-CL~\citep{gao2021contrastive}), or within group activities~\citep{bian2022self}. However, the learned space is still underestimated with diverse view representations and imbalanced sample distribution, leading to a less satisfactory clustering. 

\textcolor{black}{Recently, several studies also show that pre-training the self-supervised representation benefits the supervised~\citep{yang2021skeleton,su2021self} or semi-supervised learning~\citep{si2020adversarial} in action recognition tasks. In this paper, we investigate self-supervised representations with view invariance to improve action recognition. This is done by an end-to-end framework with balanced pairwise learning based on the performed viewpoints, such that the learned representations of the same scene are more clearly grouped with fewer errors (\textit{i.e.} higher purity and recognition accuracy).}

\subsection{View-invariant Human Action Recognition}
A robust action recognition model requires the learned representations to be less sensitive to viewpoints. \textcolor{black}{In RGB, depth, or optical flow videos, it is common for people to remove the view-dependent backgrounds by learning the view-specific focuses in different viewpoints~\citep{li2018unsupervised,wang2014cross,wang2018dividing,sun2022human}.} In contrast, a natural advantage of the 3D skeleton is that the view-invariant features are more easily extracted with the body joint positions. 
However, the commonly used view-independent representations, including the statistics-based histogram of joint orientations~\citep{xia2012view} or geometry-based pairwise joint distance~\citep{nie2019view} will discard some semantic information that is useful for recognizing action patterns. 

Another branch of works~\citep{lee2017ensemble,gao2021contrastive,paoletti2021unsupervised} employed coordinate transformation (\textit{i.e.} rotations, translations) to align or synthesize multi-view actions.
For example, \textcolor{black}{Gao et al.~\citep{gao2021contrastive} compared the action pairs augmented from arbitrary viewpoints by a contrastive loss. Paoletti et al.~\citep{paoletti2021unsupervised} utilized gradient reversing to fool a viewpoint regressor that predicts the rotation of the transformed action. 
However, with the self-occlusions in different directions, the transformed skeleton is not accurate enough to imitate the new viewpoint.} 
Moreover, Zhang et al.~\citep{zhang2019view} automatically learned the view-invariant adaption per action and achieved promising results compared with pre-defined transformations. It further convinces that the recognition ability can be dramatically affected by the inconsistency between viewpoints that cannot be removed manually. 

Recently, Gao et al.~\citep{gao2022global} exploited fisher contrastive learning to extract view semantics from different scales of body parts. 
By assembling multiple spatial features, Guan et al.~\citep{guan2022afe} proposed a feature-enhanced approach that is robust to view variation. Nie et al.~\citep{nie2020unsupervised} proposed SeBiReNet that models the view variant and invariant features from the geometric poses and temporal dynamics, respectively, to better denoise the skeleton data. Instead of specifying the view-invariant features, we purify the latent representations through the implicit correlations learned between multi-view samples, which yields a better recognition performance.

\subsection{Contrastive Learning}
Contrastive learning~\citep{wu2018unsupervised,tian2019contrastive,chen2020simple} is a self-supervised representation learning method that differentiates between individual instances based on their pairwise similarity. As a label-agnostic approach, contrastive learning is purely based on the feature-level correlations of samples, which is very popular in large-scale visual tasks. Chen et al.~\citep{chen2020simple} augmented the real-world image such as cropping, rotations, or blurring, as positive instances to extract common properties in contrastive learning, where they further used a large minibatch size to increase the capacity of contrastive learning that achieves superior classification performance. However, the above methods rely on heavy data accommodation that requires either argumentation or multi-modal representations.

There are also several works adopting contrastive learning in skeleton-based action recognition. Rao et al.~\citep{rao2021augmented} learned the self-supervised action representation with contrastive learning, where they exploited similar augmentation strategies of images~\citep{chen2020simple} onto skeleton sequences. Lin et al.~\citep{lin2020ms2l} constrained the contrastive loss with multi-task learning from motion prediction and classification. Since the learned embeddings from contrastive loss are hard to discriminate, effective selection strategies~\citep{he2020momentum, tian2019contrastive, NEURIPS2020_e025b627} are proposed for positive or negative samples to enhance the contrastive metric. As a remarkable work, Momentum Contrast (MoCo)~\citep{he2020momentum} performed contrastive learning by selecting negative samples from a memory bank that is updated dynamically with an extra momentum encoder, which is later adopted in many recent work~\citep{li20213d, thoker2021skeleton, wang2022contrast, ben2022and, guo2022contrastive} to improve skeleton-based action recognition. For example, Wang et al.~\citep{wang2022contrast} proposed a contrast-reconstruction representation network (CRRL) to contrast between the spatial postures and motion velocity to enhance the action representation learning. Li et al.~\citep{li20213d} proposed CrosSCLR to enrich the feature receptive field by contrasting augmented skeleton sequences from bone, joint, and motion features. Their model feasibility was further generalized by Guo et al.~\citep{guo2022contrastive} (AimCLR) with more diverse spatial and temporal augmentations. However, their learned representations were restricted by the augmented patterns and the cross-modality settings greatly increase the model complexity. Since the natural correlation between different camera viewpoints is still underexplored in self-supervised 3D action understanding, \textcolor{black}{in this paper, we investigate how the viewpoint affects the learned representation by contrasting multi-view information without augmenting the total data size.}

\begin{figure*}
\centering
\includegraphics[width=1\textwidth]{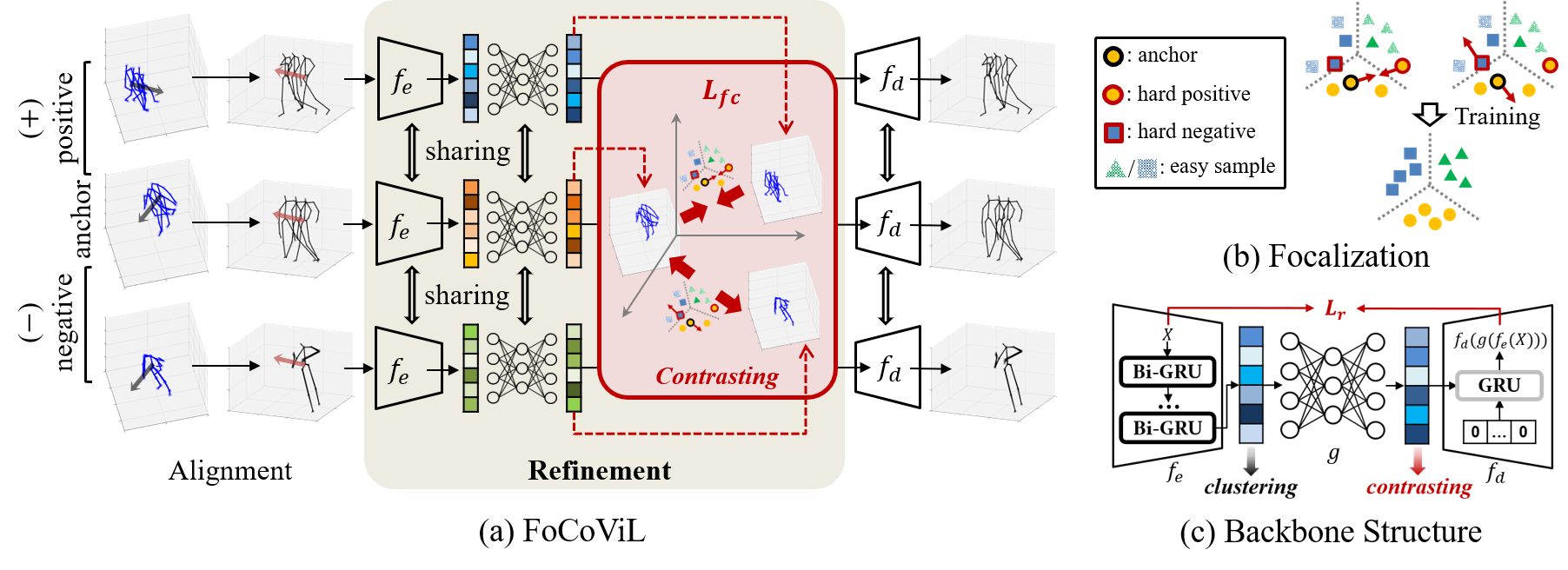}
\caption{The proposed Focalized Contrastive View-invariant Learning (FoCoViL) framework. (a) FoCoViL aims at progressively extracting the view-invariant action representations. The multi-view actions are initially aligned \textit{w.r.t.} the facing direction. In the refinement step, the proposed multi-view contrastive learning discards the implicit view-specific appearance by enlarging the agreement of the same scene under different viewpoints (``$+$" pairs), and facilitates clustering with a sparse space by enlarging different scenes (``$-$" pairs), which is further enhanced by an adaptive focalization. (b) The motivation of focalization. For each target action, focalization focuses on adjusting the hard action pairs that dominate the misclassifications. (c) The architecture of the shared auto-encoder backbone. The clustering is conducted before the encoded representation passes into the projection net $g$.}
\label{fig:framework}
\end{figure*}

\section{Methodology}
\label{sec:focovil}
We aim at progressively learning an effective representation space consisting of view-invariant action features for discriminative clustering. To achieve this goal, we propose FoCoViL, which removes the view influence on the feature-level representation extracted by an RNN-based auto-encoder backbone, as demonstrated in Fig.~\ref{fig:framework}. We first conduct a coarse-level transformation to align different viewpoints. Then, in the refinement step, we disentangle the remaining view-specific features from the latent space by finding the inherent correlations of the same action scene under different viewpoints. Since the contributions to the space learning vary from sample to sample with the {\it hard positives} (\textit{i.e.} intra-class variation) and {\it hard negatives} (\textit{i.e.} inter-class similarity) contributing more, we further propose a focalized contrastive loss to cope with the imbalanced learning complexity by adaptively adjusting their training intensity, thereby promoting the quality of the converged latent space.

\textbf{Problem Formulation} In the training set ${\bf \hat{X}}$ with $V$ viewpoints, we represent the $i^{th}$ scene of human action $\hat{X}_{i}^{u}\in{\bf \hat{X}}$ as a sequence of poses, \textit{i.e.} $\hat{X}_{i}^{u}=\{\hat{x}^{u}_{i,1}, \hat{x}^{u}_{i,2}, ..., \hat{x}^{u}_{i,T}\}$, under a specific viewpoint $u\in V$. Each pose $\hat{x}^{u}_{i,t}\in\mathbb{R}^{3\times N}$ at frame $t$ contains $N$ joint locations under 3D skeleton, and $T$ is the maximum timestamp. As a self-supervised classification task, we tend to learn a view-invariant mapping $f_e$ without the guidance of action label.

\subsection{View Alignment}
\label{sec:coarse}
The skeletons are misaligned under different viewpoints, which brings difficulty in recognizing actions. Following~\citep{su2020predict}, we transform the views for aligning the multi-view actions to the same facing direction, resulting in a more comparable representation space for the latter refinement phase. 

As 3D joint coordinates are demonstrated with different scopes under different camera points, we first translate them into a local coordinate system with the origin as the root joint $\hat{x}_{i,0}^{u}(root)$ at the initial frame, thereby removing the dependency of the camera position and the global displacement. We then match the directions of the translated actions using a rotation matrix $R=[\hat r_{0}, \hat r_{1}, \hat r_{2}]$ with: 
\begin{equation}
\begin{array}{l}
    r_{0}=\hat{x}_{i,0}^{u}(spine)\!-\!\hat{x}_{i,0}^{u}(root),\\
    r_{1}=\tilde r_{1}-\tilde r_{1}\cdot \hat r_{0},\ \textnormal{and}\ \tilde r_{1}=\hat{x}_{i,0}^{u}(lhip)\!-\!\hat{x}_{i,0}^{u}(rhip),\\
    r_{2}=r_{0}\times r_{1},
\end{array}
\end{equation}
\textcolor{black}{where $\hat x=\frac{x}{\!\parallel\! x\!\parallel\!}$ denotes the unit vector.} Here, $r_{0}$ points from the root to the spine $\hat{x}_{i,0}^{u}(spine)$, $r_{1}$ is the orthogonal projection of the vector between left $\hat{x}_{i,0}^{u}(lhip)$ and right hip $\hat{x}_{i,0}^{u}(rhip)$ on $r_{0}$.

The obtained joint $n\in N$ in the pose $x_{i,t}^{u}$ is transformed by:
\begin{equation}
    x_{i,t}^{u}(n)=R^{-1}(\hat{x}_{i,t}^{u}(n)-\hat{x}_{i,0}^{u}(root)).
\end{equation}
Therefore, the action sequence after alignment is represented by $X_{i}^{u}=\{x^{u}_{i,1}, x^{u}_{i,2}, ..., x^{u}_{i,T}\}$ within the transformed training set ${\bf X}$.

\subsection{Contrastive View-invariant Learning (CoViL)}
We propose a Contrastive View-invariant Learning (CoViL) approach to automatically refine the view-invariant features by contrasting the multi-view representations under the same and different scenes. This is to tackle the problem that many implicit view-specific appearances, such as the inferred joint positions from different directions of self-occlusions, cannot be aligned by the coarse transformation. In a closely related work~\citep{nie2020unsupervised}, the view-independent and view-dependent features are being processed as pose and temporal dependencies, respectively. However, the two types of features are not domain-specific and thus are non-trivial to be explicitly grouped. In contrast, we rely on pairwise action correlations. 
By maximizing the mutual information of the compressed representations across views, CoViL can better suppress the view-specific factors and derive a highly view-invariant latent space.

With the observation that the same scene should have closer similarity than different scenes, CoViL discards the scene-invariant (\textit{i.e.} view-variant) information 
by associating the same scene together. The objective of CoViL is to maximize the agreement of the same scene under different viewpoints (\textit{i.e.} ``$+$" positive sample pairs), as well as the disagreement of different scenes (\textit{i.e.} ``$-$" negative sample pairs) by contrastive learning. With the same action, compulsively correlating the ``$+$" pairs will reinforce the co-occurrences that are only related to the underlying action content. In addition, enlarging the differences between ``$-$" pairs will avoid an overly compact representation space while constricting the ``$+$" pairs, such that the dissimilar actions are sparsely distributed to facilitate self-supervised clustering.

Particularly, we select positive and negative pairs based on a minibatch of $I$ anchor samples that are randomly picked. For each anchor $X^u_i$ in the minibatch, we propose to increase the similarity between $X^u_i$ and its corresponding positive sample $X^v_i$ ($v\neq u$), which reduces the motion variations caused by viewpoints. Note that $X^u_i$ and $X^v_i$ are from the same scene but have different viewpoints. We also propose to maximize the dissimilarity between $X^u_i$ and its negative samples from other scenes which consist of two batches: $\{X^u_j\}_{j=1,j\neq i}^{I}$ under the same viewpoint $u$, and $\{X^v_j\}_{j=1,j\neq i}^{I}$ under the different viewpoint $v$. The far-distributed negative pairs will ensure the sparsity of the resulting space.

We integrate our proposed positive and negative pair design via the batch contrastive loss based on InfoNCE loss function~\citep{chen2020simple} due to its superior capacity in modeling the pairwise correlations. The multi-view contrastive loss defined on an anchor $X^u_i$ is given by:
\begin{equation}
	L_c(X_i^u)=-\!\!\sum\limits_{v \in V\backslash u}\!\!\log{\frac{S(X^u_i,X^v_i)}{\sum_{j\neq i}(S(X^u_i,X^u_j)+S(X^u_i,X^v_j))}},
	\label{eq:cl}
\end{equation}
where the proximity $S$ is the similarity measurement between a pair of samples. Note that the loss is summed for all the viewpoints in $V$ except $u$, and it is not bounded by only two views despite the proximity $S$ is counted based on the pairwise manner. \textcolor{black}{The proposed multi-view contrastive loss uses viewpoint information as the indicator to group positive samples under the general form of contrastive loss~\citep{chen2020simple}. Empirically, if more viewpoints are included during training, the learned representation space is more informative by disentangling multi-view features.} More specifically, $S$ between sample pair $X$ and $Y$ is determined by their encoded distance $r$ as:
\begin{equation}
S(X,Y)\!=\!\exp(\frac{g(f_e(\!X\!))\!\cdot\! g(f_e(\!Y\!))}{\tau\!\parallel\! g(f_e(\!X\!))\!\!\parallel\!\!\cdot\!\!\parallel\! g(f_e(\!Y\!))\!\!\parallel})\!=\!\exp(\frac{r(X,Y)}{\tau}),
\label{eq:sim}
\end{equation}
where $\tau$ is the temperature parameter~\citep{chen2020simple,wu2018unsupervised} that controls the scale of the similarity. We have evaluated $\tau$ with several
choices $\{0.1, 0.5, 1, 2\}$, among which 0.5 performs the best. $g$ is a projection network consisting of two fully-connected layers to integrate features, and we use the cosine distance $r$ as the similarity metric following the general form of contrastive learning~\citep{chen2020simple}. Here, the proposed multi-view contrastive loss is conducted on the projected feature 
to facilitate a more informative latent space for $f_e(X)$ that benefits the recognition.

\subsection{Focalization}
We propose to adaptively balance the easy and hard samples via dynamic focalization on the proposed CoViL, thereby increasing model robustness and reducing misclassifications. This is particularly challenging in unsupervised models, as the hard samples cannot be explicitly mined due to the lack of a label-guided distribution.

Here, we focus on rebalancing the representations within the scope of self-supervised contrastive learning, while the vast majority of other works focus on solving imbalance in supervised cross-entropy from true sample distributions. Instead of recognizing individual instances, in contrastive learning, we balance easy and hard samples via the pairwise sample similarity inherited from CoViL. As an effective solution, focal loss~\citep{lin2017focal} aims at detecting and emphasizing the hard instances from the probability outcome. While similar in purpose, our method attempts to solve the imbalanced similarity of sample pairs based on contrastive learning, from the observation that contrastive learning lacks a mechanism to maintain balanced training. By entangling and disentangling sample pairs in terms of their representation similarity, the focalized CoViL (FoCoViL) further enhances the latent space learning with a clearer decision boundary between clusters (see Fig.~\ref{fig:framework}(b)). Note that the proposed focalized contrastive learning is not a simple reweighting scheme but balancing and improving the distributions in representation space by self-supervised hard
sampling with adjustable hardness, which has more generalizable advantages and consistent performance improvements (see Table~\ref{tab:ablation}).

In particular, we propose a dynamic-scaled focal loss based on the geometric distance of the contrastive representations. Inspired by the evidence that the same scene should have similar feature expressions, we consider a ``$+$" pair as hard if they are too far distributed. Analogous to positive pairs, we define the hard negatives if the ``$-$" pair is too close. FoCoViL intuitively pulls the same scene with very different representations closer while pushing the different scenes with similar representations apart. Numerically, we monotonously increase the weight to the ``$+$" pair $X^u_i$ and $X^v_i$ if their cosine similarity $r$ is getting close to -1, and increase the weight to the ``$-$" pair if $r$ is near 1. The dynamic weight $w_{+}$ for positive pair and $w_{-}$ for negative pair are defined by:
\begin{equation}
    \begin{gathered}
    w_{+}=\sigma(1-r(X^u_i,X^v_i)),\\
w_{-}=\sigma(\frac{1}{2I\!-\!2}\!\sum_{j\neq i}[(1\!+\!r(X^u_i,X^u_j))\!+\!(1\!+\!r(X^u_i,X^v_j))]).
    \end{gathered}
\end{equation}
The modulating factors $1-r(X,Y)$ and $1+r(X,Y)$ are added as pair weightings for the positive and negative samples, respectively, which adaptively differentiate between the easy and hard pairs based on the pairwise similarity. The \textit{sigmoid} activation $\sigma(\cdot)$ with the common form $\sigma(x)=\frac{1}{1+e^{-x}}$ is to incorporate nonlinearity to contrastive loss, and the scaling term $\frac{1}{2I-2}$ is to balance the quantities of positive and negative pairs. 

By decomposing $L_c$ in Eq.~\ref{eq:cl}, the proposed focalized multi-view contrastive loss $L_{fc}$ is defined as:
\begin{multline}
L_{fc}(X_i^u)=-\!\!\sum\limits_{v\in V\backslash u}\!\![w_{+}\log{S(X^u_i,X^v_i)}\\-w_{-}\log{\sum_{j\neq i}(S(X^u_i,X^u_j)+S(X^u_i,X^v_j))}].
\end{multline}
Compared with the inliers that stay very close to the cluster centers, the hard sample pairs usually include outliers that are scattered near the cluster boundaries. By focusing on the hard pairs, we establish a more robust latent space with fewer misclassified outliers. Note that the proposed focalized contrastive loss is heuristic and can be extended to other models adopting contrastive learning.

\subsection{The Multi-view Auto-encoder Backbone}
\label{sec:autoencoder}
To maintain the action representation, we employ an effective sequential auto-encoder as the backbone network sharing among the multi-view actions (see Fig.~\ref{fig:framework}(c)). From~\citep{yang2017improved}, the encoder usually plays a more important role than the decoder to integrate representative features. We thus consider a portable decoder by feeding the empty frame (zero vector) to every step of the decoder, such that the model only focuses on the hidden representation delivered from the encoded output. Instead of a two-stage encoder for feature extraction~\citep{su2020predict}, our FoCoViL makes full use of a single auto-encoder in an end-to-end fashion that already achieves promising results.

The structure consists of a three-layer bi-directional encoder $f_e$ to derive the latent representation, a linear projection net $g$ that is specifically designed for contrastive learning, and a single-layer uni-directional decoder $f_d$ for reconstruction purposes. Both $f_e$ and $f_d$ are under the Gated Recurrent Unit (GRU) architecture to process frame-wise information. For each action $X_i^u$, the reconstruction loss $L_r$ is defined as:
\begin{equation}
	L_r(X^u_i)=\frac{1}{T}\sum_{t=1}^{T}{\parallel f_d(g(f_e(X_i^u))-X_i^u\parallel},
\end{equation}
where $T$ is the total number of frames in the action video. Since the sequential auto-encoder reconstructs $T$ frames of action, the loss is counted based on every frame and then averaged.

\subsection{Training and Classification}
The final objective of the proposed FoCoViL is given by $\alpha L_{fc}+\beta L_{r}$, where the $\alpha$ and $\beta$ are the trade-offs between two losses. By optimizing the combination of $L_{fc}$ and $L_{r}$, the whole network will search for the optimal representation space for the downstream classification task, where $f_e(X)$ is used for evaluation. Note that we do not cluster on the compressed output $g(f_e(X))$, since it may discard some information that is necessary for classification~\citep{chen2020simple}.

\section{Experiments}
\label{sec:experiments}
\subsection{Datasets and Experimental Setup}
\subsubsection{Datasets} To test the robustness of our model, we evaluate five benchmark 3D action datasets with diverse scales and properties, \textit{i.e.} N-UCLA~\citep{wang2014cross}, NTU RGB+D 60~\citep{shahroudy2016ntu}, NTU RGB+D 120~\citep{liu2019ntu}, PKU-MMD~\citep{liu2017pku}, and UWA3D~\citep{rahmani2014hopc}. The adopted datasets were all captured using Kinect with diverse self-occlusions under a multi-view environment. N-UCLA contains 10 types of human daily activities from three different viewpoints. PKU-MMD, NTU RGB+D 60, and NTU RGB+D 120 are large-scale action datasets with around 20,000, 56,880, and 114,480 clips covering 51, 60, and 120 types of human activities, respectively, where NTU RGB+D 120 is the largest benchmark for skeletal action recognition. UWA3D is more challenging due to four distinct action directions captured from the front, left, right, and top views.

\subsubsection{Implementation Details} 
For pre-processing, the raw skeleton is initially normalized to $[-1,1]$, and before feeding in the model, all action clips are interpolated to a fixed length with 50 frames. Our FoCoViL is trained under the combination of $L_{fc}$ and $L_r$, where we set $\alpha=\beta=1$ for N-UCLA, NTU RGB+D, and PKU-MMD, and $\alpha=1, \beta=2$ for UWA3D because of its noisy skeletons. $\tau$ is set to 0.5 for the similarity measurement in Eq.~\ref{eq:sim}. Inside the auto-encoder, 1024 hidden units are used in the GRU cell for each layer, and the unit sizes in the projection net are 512 and 1024 respectively. The training batch size is 128 for NTU RGB+D and 64 for the other three datasets, and we adopt Adam optimizer with a learning rate of 0.0001 and a decay rate of 0.95. Following~\citep{su2020predict} and~\citep{lin2020ms2l}, we conduct cross-view (CV) evaluations on N-UCLA, NTU RGB+D 60, and UWA3D, and cross-subject (CS) evaluations on NTU RGB+D 120 and PKU-MMD. Unlike~\citep{su2020predict} only tested two views on UWA3D, we test on all evaluation combinations with any two viewpoints for training and the rest for testing, which results in 12 experimental trials in total.

\subsubsection{\textcolor{black}{Evaluation Protocols for Classification}}
\textcolor{black}{For a fair comparison, we adopt both supervised and unsupervised classifiers to evaluate the encoded representation for the action recognition task. \textbf{Linear Classifier}: as in~\citep{li20213d,guo2022contrastive}, a fully-connected layer (together with a \textit{softmax} activation) is trained on the top of the fixed encoder as the supervised evaluator. \textbf{1-Nearest Neighbor (1-NN)}: as adopted in~\citep{su2020predict}, the test label is assigned from its top nearest neighbour in the training samples in a non-parametric fashion, where the representation similarity is measured by cosine distance. Unlike the supervised linear classifier, 1-NN is used as an unsupervised evaluator that does not require extra training to assign the label.}

\begin{table}
\caption{Performance comparisons (\%) on N-UCLA with supervised (Linear) and unsupervised (1-NN) evaluation protocol.}
\centering
\begin{tabular}{lc}
\hline
Method      & N-UCLA   \\ \hline  
\multicolumn{2}{c}{\textbf{Linear Classifier}} \\ \hline 
LongT GAN~\citep{zheng2018unsupervised}           & 74.3      \\
Denoised-LSTM~\citep{demisse2018pose}         & 76.8         \\
MS\textsuperscript{2}L~\citep{lin2020ms2l}          & 76.8  \\
SeBiReNet~\citep{nie2020unsupervised}             & 80.3     \\
AS-CAL~\citep{rao2021augmented}          & 75.6  \\
ST-CL~\citep{gao2021contrastive}          & 81.2  \\
MCAE-MP~\citep{xu2021unsupervised}              & 83.6  \\
CRRL~\citep{wang2022contrast}                                & 83.8 \\
\textbf{FoCoViL}             & \textbf{84.2}    \\ \hline
\multicolumn{2}{c}{\textbf{1-Nearest Neighbour (1-NN) Classifier}}        \\ \hline
P\&C~\citep{su2020predict}                  & 84.9     \\ 
MCAE-MP~\citep{xu2021unsupervised}              & 79.1    \\
CRRL~\citep{wang2022contrast}                                & 86.4 \\
\textbf{CoViL}               & 86.7             \\
\textbf{FoCoViL}             & \textbf{88.3}   \\ \hline
\end{tabular}
\label{tab:n-ucla}
\end{table}

\begin{table}[h!]
\caption{\textcolor{black}{Performance comparisons (\%) on NTU RGB+D 60 with supervised (Linear) and unsupervised (1-NN) evaluation protocols.}}
\centering
\begin{tabular}{lc}
\hline
Method                 & NTU RGB+D 60         \\ \hline  
\multicolumn{2}{c}{\textbf{Linear Classifier}} \\ \hline 
LongT GAN~\citep{zheng2018unsupervised}              & 48.1                   \\
SeBiReNet~\citep{nie2020unsupervised}              & 79.7            \\
AS-CAL~\citep{rao2021augmented}              & 64.8            \\
ST-CL~\citep{gao2021contrastive}              & 69.4            \\
MCAE-MP~\citep{xu2021unsupervised}              & 74.7                 \\
CRRL~\citep{wang2022contrast}                                & 73.8 \\
TSL~\citep{ben2022and}                & 76.3   \\
CrosSCLR~\citep{li20213d}               & 76.4   \\
3s-CrosSCLR~\citep{li20213d}               & 83.4   \\
AimCLR~\citep{guo2022contrastive}       & 79.7 \\
3s-AimCLR~\citep{guo2022contrastive}       & 83.8 \\
3s-SCC~\citep{yang2021skeleton}           & 83.1 \\
ISC~\citep{thoker2021skeleton}        & \textbf{85.2}\\
\textbf{FoCoViL}         & 83.2                 \\ \hline 
\multicolumn{2}{c}{\textbf{1-Nearest Neighbour (1-NN) Classifier}}        \\ \hline
P\&C~\citep{su2020predict}                   & 76.1                    \\
CrosSCLR~\citep{li20213d}               & 63.5                   \\
3s-CrosSCLR~\citep{li20213d}               &  65.2                 \\
AimCLR~\citep{guo2022contrastive}       & 70.1                \\
3s-AimCLR~\citep{guo2022contrastive}       &  69.3             \\
CRRL~\citep{wang2022contrast}                                & 75.2 \\
MCAE-MP~\citep{xu2021unsupervised}              & \textbf{82.4}            \\
\textbf{CoViL}           & 79.4                 \\
\textbf{FoCoViL}         & 80.2  \\ \hline
\end{tabular}\label{tab:ntu}
\end{table}

\subsection{Recognition Comparisons with the SOTAs}
We first compare the proposed FoCoViL with the state-of-the-art (SOTA) unsupervised approaches based on 3D skeleton, including regression-based representation learning models LongT GAN~\citep{zheng2018unsupervised}, Denoised-LSTM~\citep{demisse2018pose}, SeBiReNet~\citep{nie2020unsupervised}, P\&C~\citep{su2020predict}, MCAE-MP~\citep{xu2021unsupervised}, and SCC~\citep{yang2021skeleton}, and contrastive learning-based models MS\textsuperscript{2}L~\citep{lin2020ms2l}, AS-CAL~\citep{rao2021augmented}, ST-CL~\citep{gao2021contrastive}, CRRL~\citep{wang2022contrast}, TSL~\citep{ben2022and}, ISC~\citep{thoker2021skeleton}, CrosSCLR~\citep{li20213d}, and AimCLR~\citep{guo2022contrastive}, where action class labels of all models are not used during training. 

As shown in Table~\ref{tab:n-ucla}, our FoCoViL achieves superior recognition results on N-UCLA dataset for both evaluation protocols compared to other SOTA models, including the contrastive-based method CRRL. FoCoViL yields a significant 3.2\% accuracy increase over P\&C on N-UCLA, which shows that the disparity of the same scene from different viewpoints can greatly affect the classification results. Furthermore, we also achieve consistent improvements from CoViL to FoCoViL,
showing that focalization improves contrastive learning with better feature representation.

\begin{table}
\caption{\textcolor{black}{Performance comparisons (\%) on NTU RGB+D 120 with supervised (Linear) and unsupervised (1-NN) evaluation protocols.}}
\centering
\begin{tabular}{lc}
\hline
Method                 & NTU RGB+D 120         \\ \hline  
\multicolumn{2}{c}{\textbf{Linear Classifier}} \\ \hline 
AS-CAL~\citep{rao2021augmented}              & 48.6            \\
MCAE-MP~\citep{xu2021unsupervised}              & 52.8            \\
CRRL~\citep{wang2022contrast}                                & 56.2 \\
TSL~\citep{ben2022and}                & 59.1   \\
ISC~\citep{thoker2021skeleton}        &  67.1\\
CrosSCLR~\citep{li20213d}               & 67.1   \\
3s-CrosSCLR~\citep{li20213d}               & 67.9   \\
AimCLR~\citep{guo2022contrastive}       & 63.4 \\
3s-AimCLR~\citep{guo2022contrastive}       & \textbf{68.2} \\
\textbf{FoCoViL}         & 62.3             \\ \hline 
\multicolumn{2}{c}{\textbf{1-Nearest Neighbour (1-NN) Classifier}}        \\ \hline
P\&C~\citep{su2020predict}                   & 39.5                    \\
MCAE-MP~\citep{xu2021unsupervised}              & 42.3            \\
ISC~\citep{thoker2021skeleton}        & 50.6\\
CrosSCLR~\citep{li20213d}               & \textbf{52.5}                   \\
\textbf{FoCoViL}         & 51.0 \\ \hline
\end{tabular}\label{tab:ntu_120}
\end{table}

For NTU RGB+D 60 in Table~\ref{tab:ntu} and NTU RGB+D 120 in Table~\ref{tab:ntu_120}, the overall accuracies are slightly lower than N-UCLA for all methods since the datasets contain highly similar classes such as \textit{drinking water} and \textit{eating meal}, as well as local-scale movements such as \textit{thumb up} and \textit{thumb down}. 
\textcolor{black}{We first observe that FoCoViL is more advantageous than non-contrastive learning-based approaches like SeBiReNet and 3s-SCC. When comparing with MoCo-based approaches, FoCoViL outperforms CRRL and TSL, and performs comparably with CrosSCLR, AimCLR, and ISC under supervised evaluations.
Under the more challenging unsupervised protocol, 
FoCoViL outperforms ISC under NTU RGB+D 120, and outperforms both single- and multi-stream CrosSCLR and AimCLR under NTU RGB+D 60 with over 10\% performance improvement.
This shows that CrosSCLR and AimCLR heavily rely on the supervised classifier.} 

\begin{table}
\caption{\textcolor{black}{Comparison of model complexity based on NTU RGB+D 60.}}
\centering
\begin{tabular}{lcccc}
\hline
Method                 & CRRL~\citep{wang2022contrast} & 3s-CrosSCLR~\citep{li20213d} & 3s-AimCLR~\citep{guo2022contrastive} & FoCoViL         \\ \hline
Inference Time            & - & \textbf{0.69ms} & 0.70ms & 1.31ms \\
\#params                                & 7.4M  & 2.51M  & 2.51M  & \textbf{1.62M}                 \\ \hline
\end{tabular}\label{tab:complexity}
\end{table}

\begin{table}
\caption{\textcolor{black}{Performance comparisons (\%) on PKU-MMD with unsupervised (1-NN) evaluation protocol.}}
\centering
\begin{tabular}{lcc}
\hline
Method      & Phase 1  & Phase 2 \\ \hline
P\&C~\citep{su2020predict}                 & 70.4                      & 38.4 \\
CrosSCLR~\citep{li20213d}                  &  68.9         & 21.2 \\ 
AimCLR~\citep{guo2022contrastive}                  & 72.0          & 39.5 \\ \hline
\textbf{FoCoViL}             & \textbf{75.2}              & \textbf{43.3} \\ \hline
\end{tabular}\label{tab:pku}
\end{table}

\begin{table*}
\centering
\caption{Performance comparisons (\%) on UWA3D under different training partitions.}
\resizebox{\textwidth}{!}{
\begin{tabular}{lccccccccccccc}
\hline
Training Views         & \multicolumn{2}{c}{V1\&V2} & \multicolumn{2}{c}{V1\&V3}    & \multicolumn{2}{c}{V1\&V4} & \multicolumn{2}{c}{V2\&V3} & \multicolumn{2}{c}{V2\&V4} & \multicolumn{2}{c}{V3\&V4} & \multirow{2}{*}{\begin{tabular}[c]{@{}c@{}}Average\end{tabular}} \\
Testing Views          & V3           & V4          & V2            & V4            & V2           & V3          & V1           & V4          & V1           & V3          & V1           & V2          &                                                                         \\ \hline
AS-CAL~\citep{rao2021augmented}             & 25.1         & 22.8        & 21.3          & 19.7          & 22.4         & 25.5        & 21.6         & 19.5        & 23.9         & 21.1        & 21.2         & 19.7        & 22.0                                                                   \\
SeBiReNet~\citep{nie2020unsupervised}             & 53.9         & 61.6        & 54.1          & 58.6          & 51.5         & 52.0        & \textbf{71.5}         & 56.0        & \textbf{72.3}         & 51.3        & \textbf{68.9}         & 51.5        & 58.6                                                                    \\
P\&C~\citep{su2020predict}                  & \textbf{59.9}         & 63.1        & 57.1          & 62.7          & 58.7         & 58.3        & 63.5         & 58.3        & 64.3         & 53.8        & 66.3         & \textbf{55.2}        & 60.1                                                                    \\
\textbf{FoCoViL} & 58.3         & \textbf{63.1}        & \textbf{59.5} & \textbf{64.7} & \textbf{59.1}         & \textbf{59.9}        & 67.5         & \textbf{61.5}        & 69.8         & \textbf{56.7}        & 67.5         & 52.8        & \textbf{61.7}                                                                    \\ \hline
\end{tabular}\label{tab:uwa3d}
}
\end{table*}

\textcolor{black}{As shown in Table~\ref{tab:complexity}, we also test the computational resources by comparing the processing time for each action clip and the number of parameters used for~\citep{li20213d, guo2022contrastive} and FoCoViL, where FoCoViL has comparable classification performance but much fewer parameters. In general, FoCoViL is less spatially complex compared to other contrastive learning-based methods, specifically 3s-CrosSCLR~\citep{li20213d} and 3s-AimCLR~\citep{guo2022contrastive}, since they are multi-stream methods that fuse three skeleton features (\textit{i.e.}, joint, bone, and motion). In terms of inference time, FoCoViL takes longer. However, since the action sequence is relatively short (usually around 1s for each trial), the model complexity will not be heavily affected by the recurrent times of GRU.}

When comparing PKU-MMD in Table~\ref{tab:pku}, FoCoViL also outperforms CrosSCLR and AimCLR for both phases 1 and 2, where phase 2 is a noisy version dataset with more diversities in terms of facing directions and action performance. Note that the cross-subject evaluations are conducted on PKU-MMD, where three viewpoints are used for training that validates the feasibility of our method under multiple viewpoints ($\geq2$).

For UWA3D in Table~\ref{tab:uwa3d}, we significantly outperform AS-CAL~\citep{rao2021augmented} at all partitions of different training and testing viewpoints. \textcolor{black}{Since the UWA3D dataset is challenging with large variations appearing in different viewpoints, the performance may vary in different training combinations. However, on average our method performs better than SeBiReNet and P\&C. This is evidenced by having most of the best performances achieved using our approach, which shows the generality of our model across a variety of viewpoints.}

\begin{figure}[]
    \centering
    \subfigure[P\&C~\citep{su2020predict}]{
        \includegraphics[width=0.45\columnwidth]{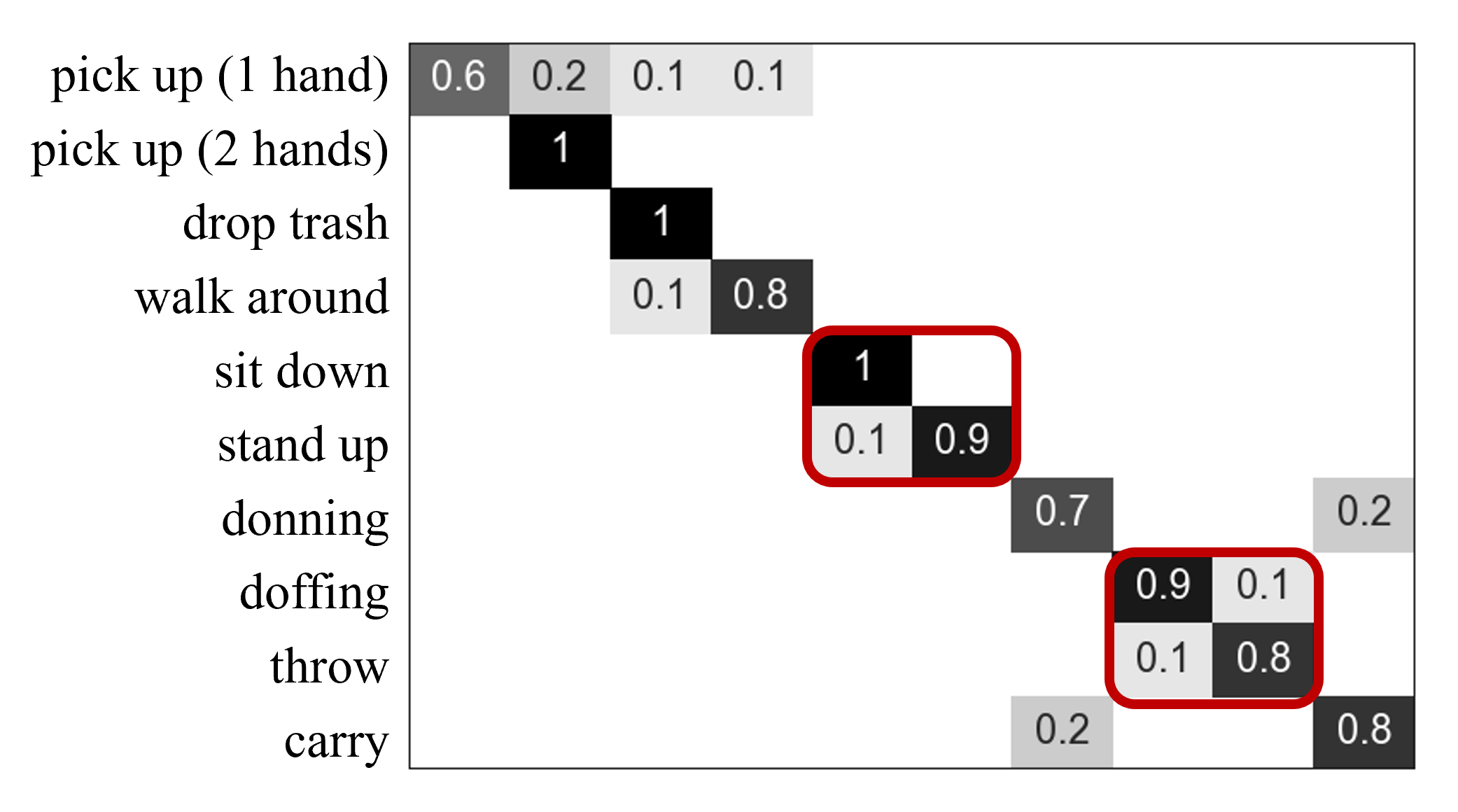}}
    \subfigure[FoCoViL]{
        \includegraphics[width=0.45\columnwidth]{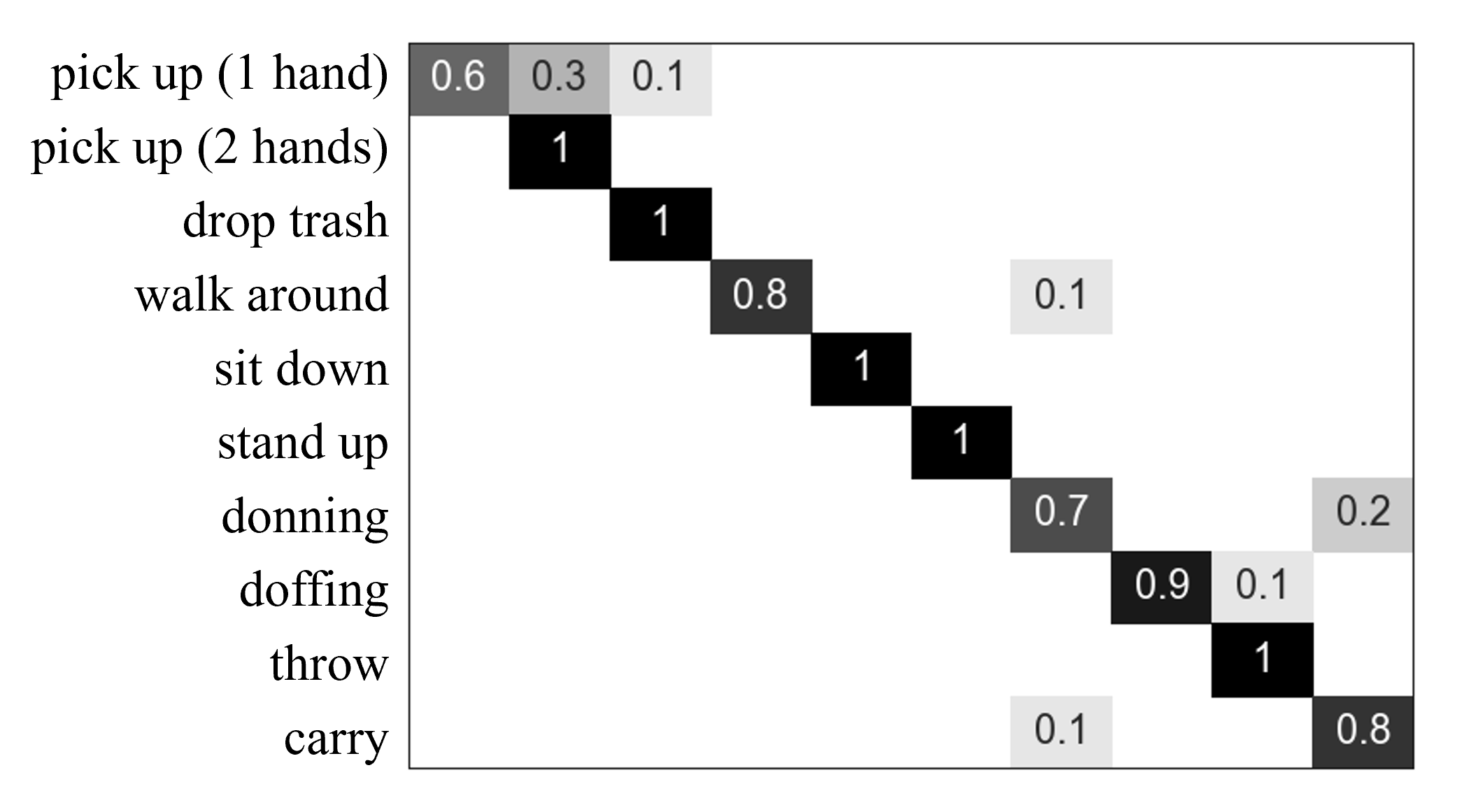}}
    \caption{Confusion matrices on N-UCLA.}
    \label{fig:cm}
\end{figure}

We further compare the confusion matrices of P\&C and FoCoViL under all types of actions of N-UCLA. In Fig.~\ref{fig:cm}, we observe that five classes reach 100\% recognition accuracy in the proposed FoCoViL compared to three in P\&C. In addition, FoCoViL discriminates better between the actions like \textit{doffing vs. throw} and \textit{donning vs. carry}. In particular, these two pairs of actions are quite similar in some viewpoints, thereby hard to be distinguished. By correlating different viewpoints, FoCoViL can get a comprehensive understanding of action features from various angles to provide a more discriminative classification for these ambiguous classes.

\begin{figure*}
\centering
    \subfigure[AimCLR]{
        \includegraphics[width=0.32\textwidth]{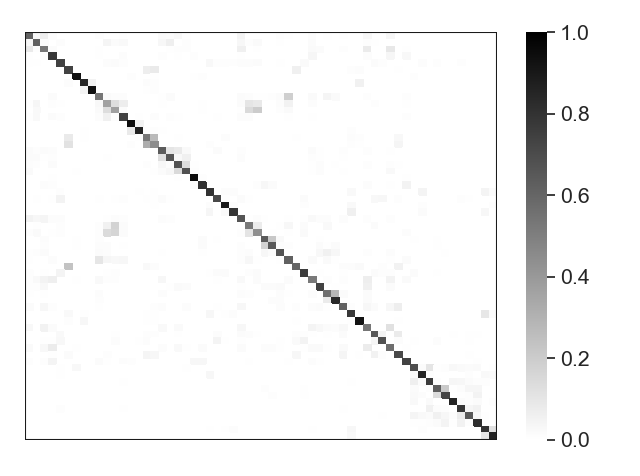}}
    \subfigure[FoCoViL]{
        \includegraphics[width=0.32\textwidth]{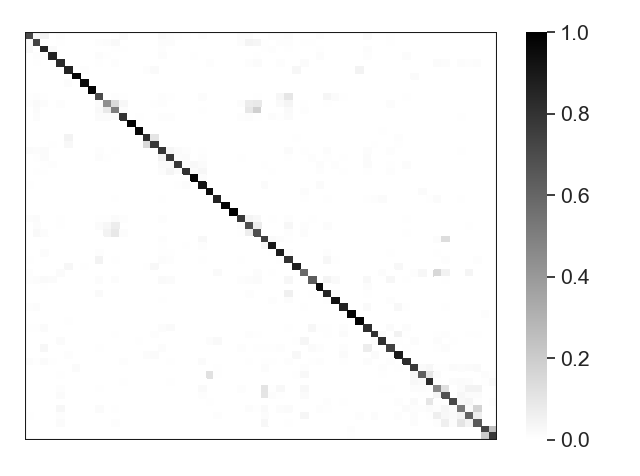}}
    \subfigure[FoCoViL-AimCLR]{
        \includegraphics[width=0.32\textwidth]{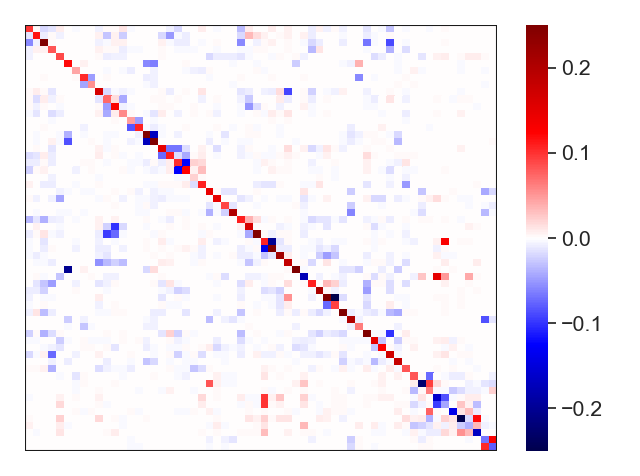}}
    \caption{\textcolor{black}{Confusion matrices on NTU RGB+D 60 on all 60 classes. (a) and (b) represent the confusion matrices of AimCLR and our FoCoViL, respectively, and (c) is the difference confusion matrix between the two methods, \textit{i.e.} (b)-(a). In (c), The diagonal value in red and the off-diagonal element in blue indicate the better classifications from FoCoViL.}} 
    \label{fig:ntu}
\end{figure*}

The confusion matrix comparison on NTU RGB+D 60 is given in Fig.~\ref{fig:ntu}. \textcolor{black}{To compare the recognition performance with large numbers of classes, we calculate the difference by subtracting the confusion matrix of the most recent method AimCLR from FoCoViL. From Fig.~\ref{fig:ntu}(c), we observe that most of the diagonal values are positive values (red), and there are lots of negative values (blue) that appear in the off-diagonal part, which indicates that compared to FoCoViL, AimCLR is more likely to have non-zero values that are misclassified as other classes.}

\subsection{Latent Space Evaluation} 
\subsubsection{Purity \& ARI}
Following~\citep{nie2021view}, we also test two common metrics, Purity and Adjusted Rand Index (ARI), to \textit{quantitatively} evaluate the quality of our learned latent space. \textcolor{black}{\textit{Purity} measures to what extent samples in a cluster belong to the true class: }
\begin{equation}
    \textcolor{black}{Purity=\frac{1}{|X|}\sum_{k}\max_{l}\omega_{kl}}
\end{equation}
\textcolor{black}{where $|X|$ is the total number of test samples. $\omega_{kl}$ is the number of samples in the $k^{th}$ predicted cluster that belongs to the $l^{th}$ ground-truth class. \textit{ARI} measures the correctness of classification concerning the mutual information between clusters: }
\begin{equation}
    \textcolor{black}{ARI=\frac{\sum_{kl}\binom{\omega_{kl}}{2}-(\sum_{k}\binom{\omega_k}{2}\sum_{l}\binom{\omega_l}{2})/\binom{|X|}{2}}{\frac{1}{2}(\sum_{k}\binom{\omega_k}{2}+\sum_{l}\binom{\omega_l}{2})-(\sum_{k}\binom{\omega_k}{2}\sum_{l}\binom{\omega_l}{2})/\binom{|X|}{2}}}
\end{equation}
\textcolor{black}{where $\omega_{k}\!=\!\sum_{l}\omega_{kl}$,  $\omega_{l}\!=\!\sum_{k}\omega_{kl}$ is the number of samples in the $k^{th}$ cluster or the $l^{th}$ class, respectively. Both measurements reveal the quality of clustering from different aspects with the maximum value of 1 if each sample gets its cluster.} In addition to the 1-nearest neighbour, we adopt two common unsupervised clustering methods, Gaussian Mixture Model (GMM) and K-Means, on the spanned latent space. The number of clusters is set the same as the number of real classes. The corresponding results are compared with SeBiReNet~\citep{nie2020unsupervised} and P\&C~\citep{su2020predict} presented in Table~\ref{tab:purity&ari}. It is worth noting that in all the compared datasets, we achieve the highest scores with significant improvements on both clustering metrics. 

\begin{table}[h!]
\centering
\caption{Quantitative evaluation of clustering quality. For both Purity and ARI, the higher value indicates better clustering.}
\begin{tabular}{llcccc}
\hline
                                             \multirow{2}{*}{Dataset} & \multirow{2}{*}{Method} & \multicolumn{2}{c}{Purity} & \multicolumn{2}{c}{ARI}         \\
                                                                             &                          & GMM  & K-Means             & GMM        & K-Means            \\ \hline
\multirow{3}{*}{N-UCLA}                                                      & SeBiReNet~\citep{nie2020unsupervised}                & 0.513    & 0.527           & 0.280          & 0.299          \\
                                                                             & P\&C~\citep{su2020predict}                     & 0.512    & 0.592           & 0.260          & 0.412          \\
                                                                             & \textbf{FoCoViL}    & \textbf{0.605}    & \textbf{0.618}  & \textbf{0.412} & \textbf{0.478} \\ \hline
\multirow{3}{*}{\begin{tabular}[c]{@{}l@{}}NTU RGB+D 60\\\end{tabular}} & SeBiReNet~\citep{nie2020unsupervised}                & 0.131    & 0.125           & 0.071          & 0.053          \\
                                                                             & P\&C~\citep{su2020predict}                     & 0.246    & 0.249           & 0.129          & 0.137          \\
                                                                             & \textbf{FoCoViL}    & \textbf{0.294}    & \textbf{0.311}  & \textbf{0.170}          & \textbf{0.172} \\ \hline
\multirow{2}{*}{PKU-MMD}                                                         & P\&C~\citep{su2020predict}                     & 0.418    & 0.409           & 0.237          & 0.237          \\
                                                                             & \textbf{FoCoViL}    & \textbf{0.483}    & \textbf{0.501}  & \textbf{0.329}          & \textbf{0.350} \\ \hline
\multirow{2}{*}{UWA3D}                                                         & P\&C~\citep{su2020predict}                     & 0.405    & 0.445           & 0.172          & 0.221          \\
                                                                             & \textbf{FoCoViL}    & \textbf{0.469}    & \textbf{0.485}  & \textbf{0.255}          & \textbf{0.272} \\ \hline
\end{tabular}
\label{tab:purity&ari}
\end{table}

\begin{figure}[t]
\centering
    \subfigure[P\&C]{
        \includegraphics[width=0.32\columnwidth]{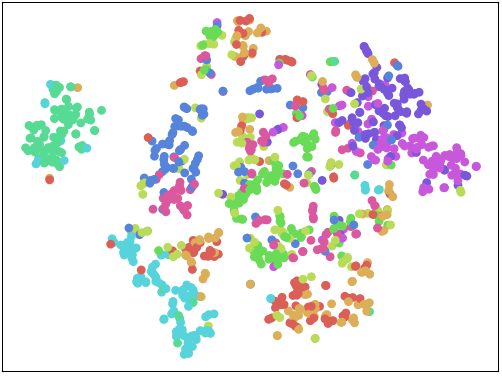}}
    \subfigure[CoViL]{
        \includegraphics[width=0.32\columnwidth]{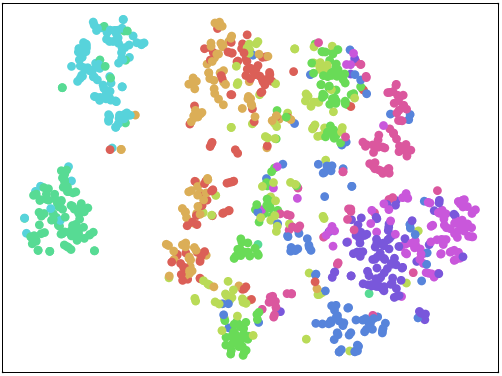}}
    \subfigure[FoCoViL]{
        \includegraphics[width=0.32\columnwidth]{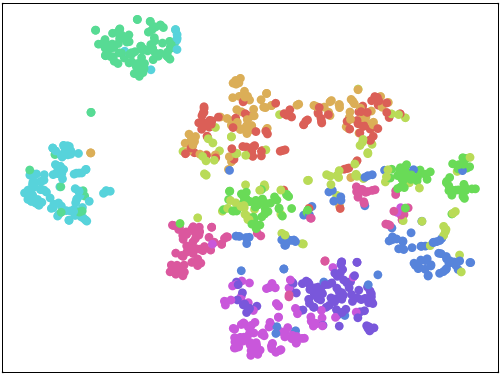}}
    \caption{T-SNE comparisons of P\&C, our CoViL, and FoCoViL on 10 classes of N-UCLA. }
    \label{fig:t-sne_ucla}
\end{figure}

\begin{figure}
\centering
    \subfigure[CrosSCLR]{
        \includegraphics[width=0.45\columnwidth]{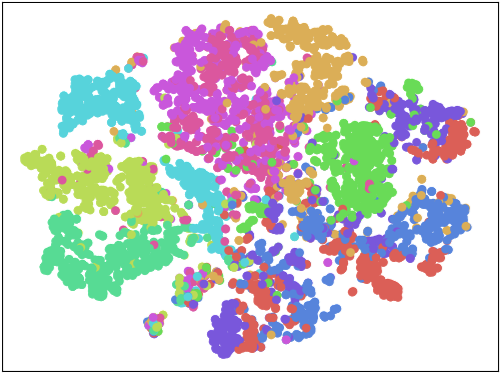}}
    \subfigure[AimCLR]{
        \includegraphics[width=0.45\columnwidth]{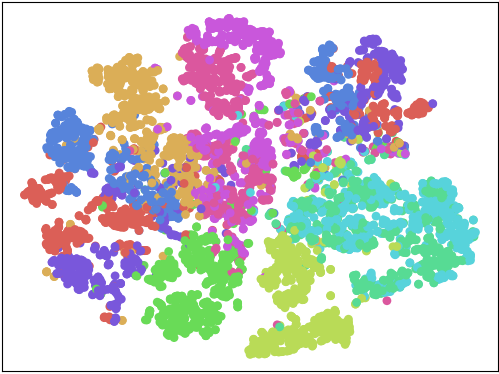}}\\
    \subfigure[CoViL]{
        \includegraphics[width=0.45\columnwidth]{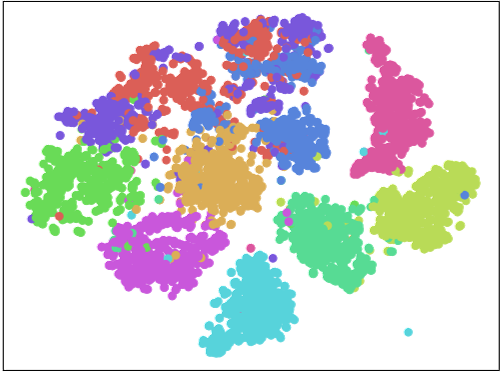}}
    \subfigure[FoCoViL]{
        \includegraphics[width=0.45\columnwidth]{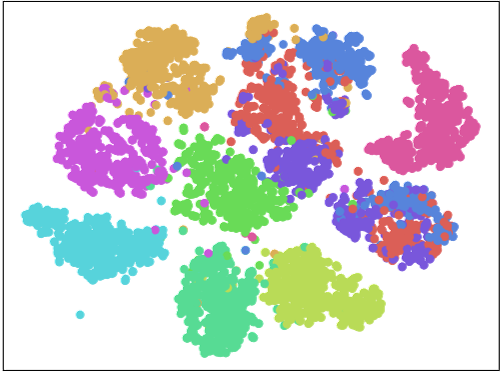}}
    \caption{\textcolor{black}{T-SNE comparisons of CrosSCLR, AimCLR, our CoViL, and FoCoViL on 10 selected categories of NTU RGB+D 60.}}
    \label{fig:t-sne_ntu}
\end{figure}

\subsubsection{Visualization}
We further visualize t-SNE of the learned features in Fig.~\ref{fig:t-sne_ucla} and~\ref{fig:t-sne_ntu} to \textit{qualitatively} compare the latent space. \textcolor{black}{The results clearly show that compared to P\&C, CrosSCLR, and AimCLR, FoCoViL generates clusters with less overlapping since it learns a more sparse and discriminative latent space by generally enlarging the distance between negative samples. Meanwhile, the representation is more compact within each cluster. This highlights that the system can better group the actions with common properties by removing the view interference.} We also visualize the effectiveness of focalization by comparing (b) and (c) of the two figures, where the latent space is improved by having a clearer margin between clusters. 

\subsection{Ablation Study} 
\subsubsection{Network Structure}
We also verify the effectiveness of the main components in our network structure. The ablation results based on the N-UCLA dataset are provided in Table~\ref{tab:ablation}. \textcolor{black}{In general, the designed modules consistently improve the performance from the reconstruction baseline (\textit{i.e.} Reconst.). We first observe that the view alignment (denoted as Ali.) can largely increase the recognition accuracy (Reconst. \textit{vs.} Ali. reconst.).} Then, by adding back the fine-level multi-view contrastive loss, the recognition performance improves by 2.9\% (Ali. reconst. \textit{vs.} CoViL), showing that the discrimination power is increased by refining the view-invariant representation. There is also a significant improvement in purity score (0.457 \textit{vs.} 0.569), proving that CoViL contributes a lot to shaping the clustering space by modeling the mutual distance between samples. Finally, by adding the focalization, FoCoViL converges to a better latent space with a clearer distribution of clusters, thus further boosting the purity score from 0.569 to 0.605.

\begin{table}
\centering
\caption{\textcolor{black}{The ablation tests of recognition accuracy and purity (GMM) on the proposed FoCoViL.}}
\begin{tabular}{lcc}
\hline
Method   & Accuracy (\%) & Purity  \\ \hline
Reconst.       & 74.4     & 0.413  \\
Ali. reconst.     & 83.8     & 0.457     \\ \hline
CoViL w/o $g$  & 84.0     & 0.531    \\
CoViL w/o ``$+$"  & 84.4     & 0.488    \\
CoViL w/o ``$-$"  & 85.1     & 0.533    \\
CoViL          & 86.7     & 0.569   \\ \hline
\textbf{FoCoViL} & \textbf{88.3}     & \textbf{0.605}      \\ \hline
\end{tabular}
\label{tab:ablation}
\end{table}

As in Table~\ref{tab:ablation}, we observe a large improvement in Purity by comparing FoCoViL (0.457 \textit{vs.} 0.605) and singly considering view alignment (0.413 \textit{vs.} 0.457). The performance gain indicates that FoCoViL contributes more to shaping the representation space to boost the clustering compared to view alignment. Although view alignment is necessary for aligning the facing directions, the resulted space is still view-dependent with large motion variations because of the view-specific self-occlusions (see Fig.~\ref{fig:view_invariant}), which explains why it is necessary to learn a view-invariant space after view alignment. By constraining the mutual distances between the same or different scenes, FoCoViL learns a better representation space with clearer cluster distributions to improve recognition.

\textcolor{black}{In addition to the main structures in FoCoViL, we also evaluate the sub-structures of the projection net $g$, CoViL w/o ``$+$" (i.e. only including negative pairs regardless of viewpoints), and CoViL w/o ``$-$" (i.e. only including positive pairs by supplementing the anchor samples from different views).} We first notice a large performance boost on both metrics by including $g$ when comparing CoViL w/o $g$ and CoViL. This extensively reflects the importance of the projection net to contrastive learning. Then we find that increasing the agreement of ``$+$" pairs (CoViL w/o ``$+$" \textit{vs.} CoViL) and the disagreement of ``$-$" pairs (CoViL w/o ``$-$" \textit{vs.} CoViL) are both essential to CoViL for a robust clustering ability, as the two factors complement each other with ``$+$" pairs generating compact clusters by correlating the same scene together, where ``$-$" pairs enable a sparse distribution to avoid an over dense representation space. 

\subsubsection{Projection Net Configuration}
As a key component of FoCoViL, we also conduct a detailed evaluation of the projection net $g$. We test the recognition performance under different combinations of layer and unit in $g$ as given in Table~\ref{tab:g}. Note that the vector dimension is 1024 for both the encoder and the decoder output. The results show that using 2 fully-connected layers by first going through a squeeze operation (1024$\rightarrow$512) in the first layer, and following an excitation operation (512$\rightarrow$1024) in the second layer will broadcast a more powerful structure to the projection net.

\begin{table}
\centering
\caption{Recognition accuracy under different structures of $g$.}
\begin{tabular}{lcc}
\hline
                          & \#input$\rightarrow$\#FC1$\rightarrow$\#FC2 & Accuracy (\%)     \\ \hline
1 Layer                   & 1024$\rightarrow$1024           & 85.1          \\ \hline
\multirow{4}{*}{2 Layers} & 1024$\rightarrow$256$\rightarrow$1024       & 85.7          \\
                          & 1024$\rightarrow$512$\rightarrow$1024       & \textbf{88.3} \\
                          & 1024$\rightarrow$1024$\rightarrow$1024      & 87.3          \\
                          & 1024$\rightarrow$2048$\rightarrow$1024      & 87.3          \\ \hline
\end{tabular}\label{tab:g}
\end{table}

\begin{table}[]
\centering
\caption{\textcolor{black}{Evaluation of training sample size.}}
\begin{tabular}{lcccc}
\hline
Data Proportion         & \textit{10\%} & \textit{50\%} & \textit{70\%} & \textit{100\%} \\ \hline
Acc. (\%) &    77.3    & 84.9   &  85.3  &   88.3    \\ \hline
\end{tabular}\label{tab:n_samples}
\end{table} 
\subsection{\textcolor{black}{The Impact of Training Size}}
\textcolor{black}{In Table~\ref{tab:n_samples}, we show how the sample size would affect the model performance by selecting 10\%, 50\%, 70\%, and 100\% (the entire training split) of data for training. When training on a small subset (10\%) of data, the performance already reaches 77.3\% with most of the actions being recognized correctly. When increasing the training proportion, the model observes more sample patterns that benefit the contrastive learning.}

\section{Conclusion}
\label{sec:con}
We propose FoCoViL for cross-view self-supervised skeleton-based action recognition in this work. By maximizing the mutual information of the multi-view actions, FoCoViL better clusters the actions with common properties in the latent space. This is done by contrasting pairwise similarity of latent representations under the same and different scenes to refine the space with high-level view-invariant features. An adaptive focalization on the contrasted sample pairs further converges FoCoViL to a more discriminative latent space with fewer misclassifications. Experiments on five benchmark 3D datasets demonstrate that our method achieves state-of-the-art recognition performance with a high-quality view-invariant space for action clustering, which has more generalization benefits. The performance also demonstrates the compatibility of FoCoViL with different scales of data size.

\textcolor{black}{Other than multi-view features, contrastive learning with focalization is also extensible to other modalities to improve the sample representation learned by contrastive loss, such as contrasting between color, depth, or textual features~\citep{tian2019contrastive,hu2021learning}.} 
Another future direction is that at the focalization stage, it is of interest to explore the imbalanced similarity of the negative pairs as well to further improve the classification performance. 

\textcolor{black}{In this work, we use an RNN-based auto-encoder to learn motion dynamics. However, the proposed focalized contrastive learning is also feasible to convolutional-based auto-encoder backbones, such as Residual 3D Convolutions (R3D)~\citep{tran2018closer} or Pseudo-3D Residual network (P3D)~\citep{qiu2017learning} that are usually adopted in RGB-based action recognition tasks, to detect more visual variations appearing in the action images.} 





\section*{Acknowledgment}
The work described in this paper was supported in part by a grant from City University of Hong Kong (Project No. 9678139) and the Royal Society (Ref: IES\textbackslash R2\textbackslash 181024 and IES\textbackslash R1\textbackslash 191147).

\bibliographystyle{elsarticle-num} 
\bibliography{ref}

\end{document}